\documentclass{article}

    \usepackage[preprint]{neurips_2025}

\usepackage[utf8]{inputenc} 
\usepackage[T1]{fontenc}    
\usepackage{hyperref}       
\usepackage{url}            
\usepackage{booktabs}       
\usepackage{amsfonts}       
\usepackage{nicefrac}       
\usepackage{microtype}      
\usepackage{xcolor}         
\usepackage{amsmath}
\usepackage{booktabs}
\usepackage{tikz}
\usepackage{array}
\usepackage{adjustbox}
\usepackage{enumitem}
\usepackage{soul}
\usetikzlibrary{positioning}   
\usepackage{verbatim}
\usepackage{cleveref}
\newcommand*\samethanks[1][\value{footnote}]{\footnotemark[#1]}

\title{Agents Require Metacognitive and Strategic Reasoning to Succeed in the Coming Labor Markets}

\author{%
    Simpson Zhang\thanks{Equal contributions.}~, Tennison Liu\samethanks~ \& Mihaela van der Schaar\\
    DAMTP, University of Cambridge \\
    Cambridge, UK \\
}

\begin{document}

\maketitle

\begin{abstract}
Current labor markets are strongly affected by the economic forces of adverse selection, moral hazard, and reputation, each of which arises due to $\textit{incomplete information}$. These economic forces will still be influential after AI agents are introduced, and thus, agents must use metacognitive and strategic reasoning to perform effectively. Metacognition is a form of $\textit{internal reasoning}$ that includes the capabilities for self-assessment, task understanding, and evaluation of strategies. Strategic reasoning is $\textit{external reasoning}$ that covers holding beliefs about other participants in the labor market (e.g., competitors, colleagues), making strategic decisions, and learning about others over time. Both types of reasoning are required by agents as they decide among the many $\textit{actions}$ they can take in labor markets, both within and outside their jobs. We discuss current research into metacognitive and strategic reasoning and the areas requiring further development.

\end{abstract}
\section{Introduction}

Significant machine learning research is currently devoted to enhancing the capabilities of AI agents in a variety of domains. However, an AI agent that is not \textit{employed} in the real world will have \textit{limited impact} in the real world. For AI agents to be meaningfully deployed, they must navigate the complex \textit{agentic labor markets} that are likely to emerge, as the efficient matching of labor supply and demand requires facilitation through a marketplace. Unlike today, where labor market decisions involving LLM agents are largely made by humans, we anticipate that future AI agents will need to autonomously manage this process due to the scale and complexity involved. Our paper examines the capabilities that individual agents must develop to succeed in such markets, with particular emphasis on \textit{metacognitive} and \textit{strategic reasoning}. We focus primarily on the supply side, where agents operate as workers, in contrast to the demand side, which will remain largely human-driven (at least initially). Nonetheless, several of our insights apply to both sides of the labor market.

Predicting future agentic labor markets is exceedingly difficult, as the introduction of AI agents is set to produce profound changes. Tremendous technological advances will be brought about by AI agents, such as enormously increased scale (greater supply of workers), higher expertise (exceeding human experts in some tasks), and faster work speed and parallelization (much faster than humans). 
However, even though changes to labor markets will be massive overall, we argue that the \textit{economic forces} that operate within current labor markets---adverse selection, moral hazard, and reputation---will persist. This is because these economic forces are extremely general and will arise in any system that includes \textit{incomplete information}, regardless of the characteristics of market participants. There will almost surely be incomplete information surrounding AI agents, including their effectiveness for different jobs and the exact actions they choose in jobs. 

\textbf{In this position paper, we argue that agents must possess both metacognitive and strategic reasoning to successfully navigate the economic dynamics of future labor markets.} Rather than enumerating an exhaustive list of required capabilities, we focus on those that most clearly distinguish this setting from other application domains. Metacognition, a form of \textit{internal reasoning}, is essential for assessing whether to accept a labor contract and determining how to execute it efficiently, requiring a nuanced understanding of the agent's own capabilities and limitations. Strategic reasoning, a form of \textit{external reasoning}, is crucial for anticipating competitors’ actions and future market demands, enabling the agent to position itself effectively for both short- and long-term success. Without these reasoning faculties, agents will be at a significant competitive disadvantage in emerging labor markets.

\begin{figure}[t]
\vspace{-2em}
\label{fig:trifecta}
  \centering
  \begin{tikzpicture}[%
      >=stealth,
      node distance = 2.2cm and 4cm, 
      every node/.style = {
        draw,
        rounded corners,
        align = left,
        inner sep = 6pt,
        font = \scriptsize
      }
    ]

    \node (forces) {\textbf{Economic Forces}\\
                    Adverse Selection\\
                    Moral Hazard\\
                    Reputation};

\node[right = of forces] (caps) {%
  \begin{tabular}{@{} l l @{}}
    \multicolumn{2}{c}{\textbf{Capabilities}}\\[2pt]
    \textbf{Metacognition}      & \textbf{Strategic Reasoning}\\
    Self-assessment      & Theory of mind\\
    Understanding Tasks      & Strategic decisions\\
    Strategy evaluation & Strategic learning\\
  \end{tabular}
};
   \path (forces) -- node[midway, draw=none, inner sep=0] (mid) {} (caps);

\node[below = 2.0cm of mid] (acts) {%
  \begin{tabular}{@{} l l @{}}
    \multicolumn{2}{c}{\textbf{Actions}}\\[2pt]
    \textbf{External Actions} & \textbf{Internal Actions}\\
    Accept jobs               & Learn about self\\
    Take effort in jobs     & Improve own abilities\\
    Hire colleagues         & Learn about others\\
  \end{tabular}
};

\draw[<->, line width=2pt, shorten >=8pt, shorten <=8pt] (forces) -- (caps);
\draw[<->, line width=2pt, shorten >=8pt, shorten <=8pt] (caps)   -- (acts);
\draw[<->, line width=2pt, shorten >=8pt, shorten <=8pt] (acts)   -- (forces);
  \end{tikzpicture}
  \caption{Trifecta view of the interactions between Economic Forces, Capabilities, and Actions.}
  \label{fig:trifecta-refined}
  \vspace{-1.5em}
\end{figure}
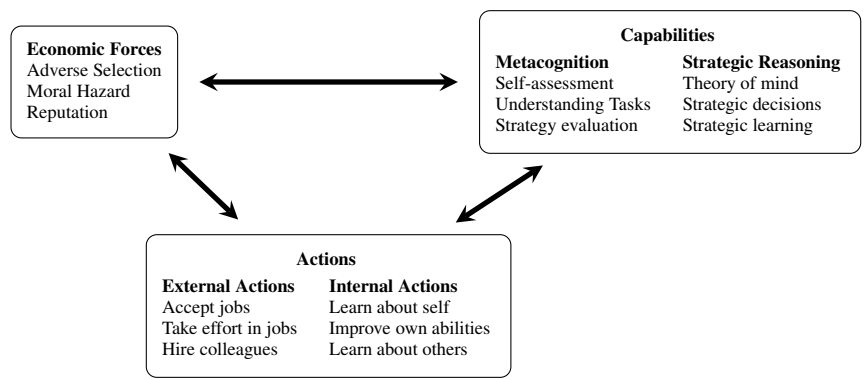

Metacognitive and strategic reasoning enable agents to determine the best actions to take within labor markets. In general, an agent must solve a complex optimization problem involving both \textit{external} actions that are job-level decisions, such as which contracts to accept and how much effort (i.e., compute and other costs) to invest, and \textit{internal} actions that are long-term self-improvement strategies that go beyond specific jobs but are essential for future competitiveness. For especially complex tasks, agents will need to reason about both the supply and demand sides of the labor market. These decisions are interconnected: an agent’s ability to hire others (demand side) may influence its decision to accept a job itself (supply side). The economic forces affecting labor markets will also evolve alongside the reasoning capabilities of AI agents, with changes to one side affecting the dynamics of the other. And the actions that the agents take in labor markets (e.g. accepting jobs, investing in skills) will evolve alongside economic forces and agentic capabilities. Figure \ref{fig:trifecta} highlights the \textit{bidirectional relationships} between economic forces, agentic reasoning capabilities, and agentic actions.

We examine the current state of agent capabilities, highlighting key areas in metacognitive and strategic reasoning where further research is needed. Labor markets present a uniquely complex and dynamic challenge, well beyond the scope of most current agent application domains. Agents capable of navigating these environments would likely generalize to a wide range of simpler settings. Beyond agent capabilities, several broader open questions arise. One concerns the design of agentic labor markets: how can they be built to ensure efficiency and fairness? Another involves the co-evolution of agents and markets, as sophisticated agents will reshape the environments they operate in, creating a feedback loop between strategic reasoning and economic dynamics. Finally, safety, alignment, and governance are critical concerns. Strategic reasoning introduces risks of misuse, manipulation, and unintended behavior, underscoring the need for robust oversight mechanisms, including auditing, interpretability, and regulation. While our focus is on the reasoning capabilities of LLM agents in labor markets, we also identify broader economic questions that warrant future investigation.

\section{Envisioning Economic Forces in Agentic Labor Markets}

\subsection{Adverse Selection and Moral Hazard}
Among the most fundamental of the economic forces present in labor markets are adverse selection, moral hazard, and reputation. Adverse selection and moral hazard are both issues that arise due to the lack of complete information, while reputation represents a critical method of providing information to a labor market. We note that adverse selection and moral hazard most directly affect the \textit{demand} side, while reputation most directly affects the \textit{supply} side of the labor market. However, given that supply and demand are jointly determined, any force that impacts one side will also affect the other.

\textbf{Adverse selection} arises when there is incomplete information about the quality or abilities of participants in the labor market. For instance, the coding ability of a new software engineer or the artistic ability of a burgeoning painter may be uncertain, making it difficult for the labor market to accurately value and compensate these individuals. Employers need to consider adverse selection when offering employment contracts, and may need to structure contracts to \textit{screen} out low-quality workers \citep{lester2019screening}. Adverse selection will have a large impact on AI agents as well, especially when an AI agent is first introduced and public experience with it is limited. Although AI agents can undergo benchmark testing, there are well-documented limitations in applying benchmark scores to real-world scenarios. Adverse selection may lead to slower uptake of AI agents, reduced willingness to pay, and fewer opportunities for AI agents to gain and learn from real-world experiences. 

\textbf{Moral hazard} arises when agents can take hidden actions that are not in the best interests of their clients \citep{holmstrom1979moral}. For instance, a lawyer may overcharge their clients on billable hours due to the client’s inability to monitor them, or a researcher may falsify data to magnify the importance of their findings. Employers need to consider potential moral hazard issues when offering contracts, as the contracts must be designed to incentivize agents to take their preferred actions. AI Agents can suffer deeply from moral hazard concerns as well. These issues could be due to alignment problems during their training process that result in unethical behavior, or they could emerge even if the agent is merely trained to maximize the profits of its principal AI company. Many current AI contracts already use a pay-as-you-go model based on the number of tokens consumed. As such, AI agents have an incentive to generate artificially long responses to charge their clients more money. This concern would be exacerbated with AI agents, as much of their work is “invisible” (e.g., manipulating background files) and would thus be more difficult to trace and oversee. 

\textbf{Mathematical Model.} We formalize adverse selection and moral hazard with a mathematical model.  
Consider a setting where an agent~\(i\) can be hired to complete a one‑time job for a client.  
The agent has a (multidimensional) type \(\theta_i\in\mathbb{R}^{N}\) that affects its productivity on the job, as well as the choice to take a costly (multidimensional) action \(a_i\in\mathbb{R}^{M}\) that increases the job output but imposes a cost on the agent of \(c(\theta_i,a_i)\in\mathbb{R}\). We denote the reward function, which represents the economic value generated by the job, as \(Z(\theta_i,a_i)\in\mathbb{R}\).
The function is increasing in both arguments: higher type implies a more capable agent, and higher action implies more effort.  
Importantly, \(Z(\theta_i,a_i)\) includes randomness, either due to factors outside the agent’s control or imperfect information about the job.

The agent makes two choices in this model.  
First, it either accepts or rejects a contract from the client.  
Second, conditional on accepting, it chooses an action \(a_i\).  
The client observes neither the agent’s type \(\theta_i\) nor its action \(a_i\), so the contract the client offers cannot be conditioned on them.  
The client’s inability to observe \(\theta_i\) captures \textit{adverse selection}, and the inability to observe \(a_i\) captures \textit{moral hazard}.  
The payment specified in the contract can therefore depend only on the realized reward; denoted by the function \(p\!\bigl(Z(\theta_i,a_i)\bigr)\).\footnote{ \(p\!\bigl(Z(\theta_i,a_i)\bigr)\) is, in general, set through bargaining between the client and the agent. As such, agents would also have to reason about this bargaining step. We abstract away from bargaining to focus on other issues.}  
Because \(Z(\theta_i,a_i)\) is random, the payment is also random. The client’s profit when hiring the agent is
\(
\Pi(\theta_i,a_i)=Z(\theta_i,a_i)-p\!\bigl(Z(\theta_i,a_i)\bigr).
\) The agent’s payoff from accepting the contract and taking action \(a_i\) is
\(
U_i(\theta_i,a_i)=p\!\bigl(Z(\theta_i,a_i)\bigr)-c(\theta_i,a_i).
\) Both client and agent have an outside option of \(0\), so they will contract only if their expected profit is positive.  

\subsection{Reputation}
In general, reputation is a store of information about past actions by an agent. If an agent performs well at a previous job, their reputation increases, and vice versa \citep{holmstrom1999managerial}. Reputation is considered a \textit{disciplining force} on agent behavior, and a \textit{helpful force} for employers to alleviate moral hazard and adverse selection.  Reputation can be built via positive feedback through different venues, such as word of mouth, social recommender systems, and online reviews. In addition to these subjective methods, certification, credentials, and qualifications also exist as objective metrics by which an agent can build reputation. Reputation systems already exist for modern LLM agents, e.g., based on benchmarks and word-of-mouth impressions (by \textit{vibes}), stratifying into frontier models vs. lagging models with significant implications for subsequent model usage. AI benchmark scores serve as a form of certification, although the correlation between benchmark scores and real-world performance can often be tenuous. These modes of developing reputation will exist for AI agents as well, but increasingly the actual job performance of an AI agent will become more highly valued. 

\textbf{Mathematical Model of Reputation.} To incorporate reputation, we add a time dimension to the above model by allowing the agent to work across multiple periods. In
each period, the agent can also work many jobs simultaneously, since agents can be
replicated (the main limitation being compute cost). The profit function of a client that hires the agent in period \(t\) for a job \(k\) as
$
\Pi_t^{\,k}\bigl(\theta_{i,t}, a_{i,t}\bigr)
    = Z_t^{\,k}\bigl(\theta_{i,t}, a_{i,t}\bigr)
      \;-\;
      p_t^{\,k}\!\Bigl(Z_t^{\,k}\bigl(\theta_{i,t}, a_{i,t}\bigr)\Bigr).
$ Note that the agent’s type \(\theta_{i,t}\) and action \(\smash{a_{i,t}}\) were already
multidimensional in the last section.  In this multi‑period, multi‑job model, the type can be
reinterpreted to include all aspects of quality across all the agent’s jobs, and the action
represents all actions the agent takes across all its jobs. The profit of the agent across all their jobs in the period \(t\) is\footnote{While payments across different jobs are additive, the costs are not in general, as the marginal cost of adding compute power is increasing due to real-world shortages in compute resources.}
\(
U_{i,t}\bigl(\theta_{i,t}, a_{i,t}\bigr)
  = \sum_k p_t^{\,k}\Bigl(Z_t^{\,k}\bigl(\theta_{i,t}, a_{i,t}\bigr)\Bigr)
    \;-\;
    c_t\bigl(\theta_{i,t}, a_{i,t}\bigr).
\) At time \(t = 0\), the expected discounted profit of the agent is
\(
\sum_{t=0}^{\infty} \delta^{t}\,
      U_{i,t}\bigl(\theta_{i,t}, a_{i,t}\bigr),
\)
where \(\delta <1\) is the agent’s discount factor. This is the objective function that the agent maximizes in an infinite‑period model.

Now that we have extended the model to multiple time periods, we can mathematically introduce the
economic force of reputation.  We assume that there is a history \(H_{i,t}\) that is influenced by
the actions taken by the agent in previous periods.  This history can be quite general and may
contain information such as the precise outputs in previous periods
\(Z_{t-1}^{\,k}\bigl(\theta_{i,t-1}, a_{i,t-1}\bigr)\), reviews left by previous clients, benchmark
scores on tests, or certifications the agent has achieved.  For a given history, \(H_{i,t}\) the
reputation is defined as the updated belief about the agent’s type.  Denote the multidimensional
posterior probability distribution of the agent’s type, conditional on the history, by
\(
f\bigl(\theta_{i,t} \mid H_{i,t}\bigr).
\)

Reputation is critical because it affects how much a client will be willing to pay an agent for
a job.  In general, an agent with a higher reputation is more likely to produce higher output in
each period and is therefore more valuable to a client.  A prospective client who sees a strong
reputation will be willing to offer a contract
\(p_t^{\,k}\!\bigl(Z_t^{\,k}(\theta_{i,t}, a_{i,t})\bigr)\) that pays the agent more because it
expects to receive higher output. Agents can undertake different jobs to improve their reputation, as well as participate in benchmark
testing or achieve certifications.  
Taking more tasks can be beneficial for an agent's reputation, as more tasks allow more signals, such as client reviews, to be recorded in
its work history.

\section{Metacognitive and Strategic Reasoning}

\subsection{Overview of Model and Agent Choices}
In the previous section, we introduced a mathematical model that covered the economic forces that underlie labor markets. In this section, we build on that model by adding AI agent metacognitive and strategic reasoning about these forces. The objective of the agent is to maximize its expected discounted profits given the costs. We expand the set of actions that the agent can take in several ways to account for metacognitive and strategic reasoning. Importantly, in addition to the \textit{external actions} $a_{i,t}$ related to the agent's jobs that we used above, we also include \textit{internal actions} $a'_{i,t}$ that are taken privately by the agent. External actions are taken within a job and directly affect the rewards from a job, whereas internal actions do not affect this period's jobs, but are instead related to self-learning and self-improvement and are useful for future periods. As such, external actions can be thought of as \textit{short-term} actions, while internal actions can be thought of as \textit{long-term} actions. Each action must be taken in consideration of the economic forces discussed above.

\textbf{Optimization Problem.} The agent’s optimization problem is given by the following:
\begin{equation}
  \max_{\{a_{i,t},a'_{i,t}\}}
  \sum_{t=0}^{\infty}\delta^{t}
    \Bigl[
      \sum_{k}
        p_{t}^{k}\!\bigl(Z_{t}^{k}(\theta_{i,t},a_{i,t})\bigr)
      -c_{t}(\theta_{i,t},a_{i,t},a'_{i,t})
    \Bigr]
\end{equation}

\textbf{Types of Actions.} The \textit{external actions} $a_{i,t}$ include: (i) Accept (or not) its job offers, (ii) Task-solving strategies and effort put into jobs it accepts, (iii) Hiring decisions and payments for colleagues. The \textit{internal actions} $a'_{i,t}$ include: (i) improving self-assessment (metacognition), (ii) improving task related capabilities (self-improvement), (iii) improving strategic decision-making. 

\textbf{Types of costs.} There are both \textit{explicit costs} that are captured directly in the cost function $c_{t}(\theta_{i,t},a_{i,t},a'_{i,t})$ and \textit{implicit costs} such as opportunity costs and reputation costs. Explicit costs are costs that the agent directly bears through spending compute resources, such as effort spent on a task or compute spent on self-improvement. By contrast, implicit costs are not part of the cost function itself. Opportunity costs are the losses in potential payments from an agent not taking a job, and reputation costs are the damages to future income caused by a loss in reputation.

\subsection{Metacognition and Self-Improvement}

Metacognition refers to the ability of an agent to learn about and adapt its own capabilities.  
There are three components of metacognition: understanding of self, understanding of tasks, and understanding of strategies to solve tasks.  
By including metacognition into the model, we can expand the agent’s action space in several ways to capture its capacity for learning about own type, learning about tasks, and learning about strategies. Collectively, we refer to these three types of actions as \textit{self-learning}.

\textbf{Improving self-assessment.} We assume that the agent does not perfectly know its own type \( \theta_{i,t} \), but it can learn about this type by performing a series of costly internal experiments at each time step. Each experiment reveals a signal \( s_t \sim g(\theta_{i,t}) \), which is potentially multidimensional, \( s_t \in \mathbb{R}^J \), and the cost of the signal is given by \( c(s_t) \), which is random and must be inferred. The agent must weigh the costs of the signal against the benefits of additional information. Based on these experiments, the agent updates its beliefs about its quality, which we denote as its \textit{self‑esteem}
\(
S_t \;=\; f\!\bigl(\theta_{i,t}\mid s_t,\, SH_{t-1}\bigr).
\)
In general, an agent’s self‑esteem is more accurate than its reputation, because reputation is based solely on publicly available information, whereas self‑esteem also incorporates the agent’s private information, denoted as the \textit{self-history}, \( SH_t \). The agent can also learn about its type through other methods, such as taking on real‑world jobs. Importantly, information gleaned from internal experiments can be of a fundamentally different nature from real‑world experience obtained through jobs, so that some aspects of the agent’s type \( \theta_{i,t} \) cannot be revealed without real‑world experience.

\textbf{Learning about tasks.} We assume that the agent can perform another set of costly internal experiments that reduce the random components of the production functions $\{Z_t^k\}$. For instance, there could be a random noise term $\varepsilon_t^k$ that affects the reward from the task, independently of the agent's type and action. By performing the action to learn about the task, the agent reduces the variance of the $\varepsilon_t^k$ term, thereby making the output more deterministic. This can be thought of as reducing the epistemic uncertainty present in the agent's model of the task's reward function.

\textbf{Learning about strategies.} We assume that the agent can take a costly action which increases the dimensionality of its external action space $a_{i,t}$. This allows the agent to take additional on-the-job actions that were not known to it before, with the new actions representing new job strategies that the agent has discovered through metacognition. For instance, the agent could discover that using another programming language allows for more efficient code. In general, these new actions may or may not achieve higher rewards than the old actions, and so the agent would still need to experiment further to learn their exact efficacy. The agent's type $\theta_{i,t}$ will also affect the rewards from these new actions. As such, if the agent's type is low, then the rewards from these new actions may also be low.

\textbf{Self-improvement.} Agents can also engage in \textit{self‑improvement}, an essential activity that requires metacognition.  
The agent’s future type \( \theta_{i,t+1} \) is affected by its current actions \( a_{i,t} \), some of which are explicit self‑improvement actions (e.g.\ internal training), according to an improvement function \( h \):
\(
\theta_{i,t+1} \;=\; h_{i,t}\!\bigl(\theta_{i,t},\, a_{i,t}\bigr).
\) The function \( h \) may include random components, and it also has a random cost.  
An agent must thus choose an action under this uncertainty. Self‑improvement can arise both from internal training and from actions performed in the agent’s jobs. This captures the fact that the agent can improve autonomously and via ``on‑the‑job training''. Crucially, some kinds of improvement are only possible through real job experience and cannot be achieved internally. Hence, the agent may choose low‑paying jobs if the experiential learning is valuable in the long run.

\subsection{Strategic Reasoning}
Now we consider the agent’s use of strategic reasoning in the labor market. Strategic reasoning includes as components strategic decision-making, strategic-learning, and theory of mind. The first component, strategic decision-making, allows the agent to make decisions in the presence of competing agents. To model this, we extend the model by adding competition with and learning about other agents. We also discuss the ability of agents to hire other agents as colleagues in the appendix.

\textbf{Model of competition.} For each job, the agent’s output function is no longer simply a function of its own actions; it also includes the actions of potential competitors, which may be a combination of humans and agents. The reward function for job \(k\), \(Z_t^{k}\!\bigl(\theta_i,a_i,\{\theta_{-i},a_{-i}\}\bigr)\), depends on the types and actions of the agent’s competitors. In general, this reward function is increasing in the agent’s own type and action and decreasing in the competitors’ types and actions.
Crucially, the agent is unable to observe the types or actions of the other agents directly due to adverse selection and moral hazard. As such, it must make inferences based on the other agents' reputations, which requires \textit{strategic reasoning}. The agent may not even know the full set of its competitors, and so it must reason about which competitors will choose to participate in the market overall and in each available job.

In general, strategic decision-making involves solving a computationally challenging problem, especially when many competitors with hard‑to‑evaluate reputations are present. There is also a challenging, dynamic aspect to this problem: the number of competitors can change over time as new competitors enter.  For example, an agent might be confident that it is currently the most capable system (i.e.\ it has the highest \(\theta_{i,t}\)), yet in the next period a more capable AI agent with a much higher type could be developed. The agent must take this possibility into account as it reasons in the current period about which jobs to accept, which actions to take in each job, what internal experiments to run, what self‑improvement investments to make, and how it expects competitors to behave.

\textbf{Other strategic reasoning components.} To model the other two components of strategic reasoning (strategic-learning and theory of mind) we allow the agent to take costly actions that provide new information about the other agents in the market. The signals received by the agent could be regarding other agents' types, past effort levels, and internal actions taken. Crucially, this inference goes beyond publicly available reputations, thus giving the agent a competitive advantage. This process captures the agent using compute resources to reason about both the beliefs (theory of mind) and the strategies (strategic learning) of other agents, thereby coming to more precise estimates of both factors.
\section{Towards Metacognitive and Strategic Reasoning in LLM Agents}
\label{sec:current_state}

Prior works have examined the economic implications of LLM agents \citep{eloundou2024gpts}, their use in specific economic applications \citep{guo2024large}, and the importance of metacognition and strategic reasoning in particular domains \citep{didolkar2024metacognitive, zhangllm}. However, the novelty of our work lies in formally envisioning future agentic labor markets, their economic forces, and the core capabilities required in such environments. Rather than contrasting with related positions, we conduct a critical assessment of current capabilities, highlighting key gaps that must be addressed.

\subsection{Metacognition}
\label{subsec:current_state_metacognition}

Metacognition allows an LLM to introspect and calibrate self-assessment of capabilities, evaluate task characteristics in a marketplace, and choose suitable strategies to complete each task.

\textbf{Self-assessment.}~Current investigations reveal that LLM agents exhibit partial self-knowledge, particularly in controlled settings. These include the ability to identify their own knowledge boundaries \citep{prato2025large} and demonstrate reasonable calibration when estimating the correctness of their outputs \citep{kadavath2022language}. However, performance on novel or out-of-distribution tasks remains weak, underscoring the difficulty of robust generalization. One of the most significant challenges to metacognitive self-assessment arises from hallucinations—failures to accurately represent internal knowledge states \citep{bender2021dangers}—as well as unreliable self-evaluations and goal awareness \citep{li2024think}. Interestingly, evidence suggests that larger model architectures tend to show improved confidence calibration and greater metacognitive sensitivity \citep{steyvers2025metacognition}, implying that some aspects of self-assessment may be learnable. While intrinsic self-calibration remains limited, external scaffolding, such as instructional prompting \citep{zhou2025calibrating} and analysis of output token likelihoods \citep{steyvers2025metacognition}, can substantially enhance metacognitive behavior. 

\textbf{Task understanding.}~An agent’s ability to comprehend the nature, demands, and difficulty of a task is critical for selecting appropriate strategies and allocating resources effectively. Evidence suggests that models possess some awareness of task requirements relative to their own capabilities \citep{li2025adaptive}, and can gauge task difficulty when explicitly prompted \citep{kadavath2022language}. However, this understanding often appears shallow; for instance, relying more on surface-level cues than on deeper semantic reasoning \citep{haroon2025accurately}. Their sensitivity to prompt structure, sometimes referred to as ``instructional distractions'', further underscores this fragility \citep{hwang2025llms, errica2024did}. Task understanding also appears to deteriorate under ambiguous instructions, with models often hallucinating rather than seeking clarification \citep{kuhn2022clam}. Task understanding also plays a critical role in long-term learning: accurate task assessment enables agents to identify self-improvement opportunities. For example, \citet{sachdeva2024train} found that LLMs can directly identify useful training samples, while \citet{wang2023voyager, wu2024copilot} showed that agents can leverage self-awareness to acquire new skills in an open-ended, self-improving manner.

\textbf{Evaluation of strategies.}~When faced with a task, modern LLM agents can choose among various problem-solving strategies, each with different cost-performance trade-offs: for instance, engaging in costly long-form reasoning or tool-use, or opting for cheaper alternatives. The ability to select and apply the most suitable strategy is therefore critical. Frameworks like ReAct \citep{yao2023react} allow agents to dynamically adjust their approach, such as retrieving information when uncertain. \citet{wu2025more} showed that optimal reasoning strategies exist for different tasks, though LLMs often struggle to determine appropriate depth. Other work has found that agents can classify mathematical problems by solution strategy \citep{didolkar2024metacognitive}, and that self-evaluation techniques, enabled by specialized prompting, can enhance strategy selection and performance \citep{zheng2024take}.

Current LLM agents exhibit early signs of metacognition, but these remain fragile, task-specific, and hard to scale. Competing in agentic labor markets will require more robust, generalized abilities—enabling agents to continuously assess their skills, adapt to shifting task demands, and identify self-improvement opportunities. Without such capabilities, they will struggle to optimize both job-level and long-term actions in response to economic pressures, limiting their competitiveness.

\subsection{Strategic Reasoning}
\label{subsec:current_state_strategic}

Strategic reasoning is an agent's extrospective ability to understand, anticipate, and adapt to others' behavior within the interactive, competitive dynamics of future agentic marketplaces.

\textbf{Theory of mind (ToM).}~A core requirement in such environments is the ability to model the beliefs, intentions, and capabilities of other agents, forming the basis for strategic reasoning. Recent studies using standardized benchmarks show that LLM agents can perform at levels comparable to human children \citep{van2023theory}, and even matching adult performance \citep{strachan2024testing}. Intriguingly, mechanistic interpretability research has identified internal representations of both self and others’ belief states in current LLMs \citep{zhu2024language}, suggesting that rudimentary ToM-like processing may emerge from pretraining. Specialized prompting and finetuning techniques have shown to further improve ToM performance \citep{sclar2024explore,sarangi2025decompose}, yet, agents continue to struggle with nuanced ToM scenarios, indicating that their grasp of social dynamics may be superficial rather than grounded in causal understanding \citep{shapira2024clever,nguyen2025survey}.

\textbf{Strategic decision-making.}~Building on ToM, strategic decision-making responds to those beliefs, by calculating optimal decisions given expectations about other agents. Existing research have investigated agent's static reasoning to achieve Nash Equilibrium in game-theoretic settings, demonstrating strategic thinking on stylized matrix games \citep{gandhi2023strategic} (e.g., ``Prisoner's Dilemma'' \citep{akata2025playing}, ``Guess the Number'' \citep{dagaev2024strategizing}) although they frequently deviate from rational strategies with larger payoff matrices \citep{hua2024game} or settings requiring long-horizon sequential reasoning \citep{duan2024gtbench}. Empirical evaluations highlight that models often rely on superficial cues rather than genuine game-theoretic reasoning, exhibiting only coarse sensitivity to different game structures \citep{lore2024strategicbehavior}. This is also notable in the paradoxical finding that agents demonstrate more sophisticated strategic behavior in complex, narrative-rich games (e.g., Werewolf \citep{bailis2024werewolf}) than complete information games \citep{topsakal2024benchmarking}, suggesting strategizing relies on language understanding and world knowledge rather than rigorous, strategic calculations. However, when augmented with game-theoretic planner and opponent modeling, LLM-based agents can achieve much stronger strategic performance, for instance attaining human-level performance in Diplomacy \citep{meta2022human}. 
Additional concerns have been raised where strategic reasoning exhibit behavioral biases (e.g., role-playing as female and heterosexual identities results in better performance), underscoring the need for further alignment \citep{jia2025evaluation}.

\textbf{Strategic learning.}~There is also a question regarding the adaptive component, whether strategic reasoning can be updated based on observations in past interactions. In this regard, LLM agents have shown capacity to adapt in response to feedback on past observations, including in the context of repeated games \citep{akata2025playing} and settings with under-defined utilities (including multi-agent simulations  \citep{park2023generative,li2023tradinggpt} and multi-agent debates \citep{taubenfeld2024systematic}). Other works have investigated using imitation or reinforcement learning to improve strategic reasoning capabilities, with notable successes in narrow two-player chess domains \citep{feng2023chessgpt,wan2024alphazerolike}. Settings like the agentic marketplace, where multiple agents interact under partial observability and optimize long-term utilities dynamically, are a central focus of multi-agent reinforcement learning (MARL) \citep{littman1994markov}. Recent progress has largely demonstrated scalable learning in complex, but narrower, game domains (e.g., Starcraft \citep{vinyals2019grandmaster}, Diplomacy \citep{meta2022human}). 

While early signs of strategic reasoning are evident, current capabilities remain limited in complexity, generality, and adaptability. In dynamic agentic economies, success will depend not only on task performance but also the ability to anticipate, compete/coordinate, and learn from interactions over time. Robust strategic reasoning, grounded in theory of mind, dynamic decision-making, and adaptive learning, will be essential for navigating economic markets and optimizing long-term utility.

\section{Open Questions}

\subsection{The Way Forward: Cross-Disciplinary Collaboration}

The introduction of AI agents has the potential to upend current systems and create massive systemic changes. To assess these changes, it is necessary to develop new modes of research in both the economics and machine learning fields, as well as cross-disciplinary research. Traditional economic models, while powerful, are often too stylized to capture the full nuance of real-world complexities and agent interactions \citep{simon1955behavioral}. The standard assumption of perfect rationality in economics may not adequately describe the decision-making of AI agents, which, though sophisticated, operate on complex, learned, and not necessarily classically rational foundations. Conversely, machine learning research can derive significant benefits from a deeper integration of economic concepts to direct the progression of agent reasoning towards real-world efficacy. For instance, AI agents that lack robust strategic reasoning capabilities to manage their reputation or navigate complex incentive structures will likely falter in competitive marketplaces \citep{parkes2015economic}. Furthermore, the development of AI agents can be greatly enhanced through collaboration in building strategically complex simulated environments for training advanced capabilities \citep{rahwan2019machine}.

\subsection{Developing Metacognition and Strategic Reasoning}

Our analysis in \Cref{sec:current_state} shows that current LLM agents display early signs of metacognitive and strategic reasoning—some emerging from pretraining scale, others enabled through external scaffolding such as prompt engineering, retrieval-augmented generation, and in-context learning. A central question is whether these approaches are sufficient for agents expected to operate autonomously over extended periods in dynamic markets. Reliance on external scaffolding risks obsolescence as market conditions dynamically evolve. This vulnerability arises from two core issues: the brittleness of scaffolding, which often fails in unfamiliar scenarios, and its limited scalability, as designing robust scaffolds for every contingency in complex markets becomes increasingly impractical.

A more promising path lies in developing metacognitive and strategic reasoning as intrinsic capabilities of the model, rather than relying on externally imposed scaffolding. While some of these skills may emerge through large-scale pretraining, this approach faces diminishing returns, and it remains unclear whether next-token prediction alone can reliably instill metacognitive insight or strategic acumen \citep{schaeffer2023emergent,du2024understanding}. Learning from demonstrations or expert trajectories \citep{lightman2023let} is another avenue, but it often depends on costly and limited human annotations. A more scalable alternative is to train agents in simulators or sandbox environments \citep{costarelli2024gamebench,wang2024tmgbench}, where they can receive continuous feedback—from market outcomes, user interactions, or internal evaluations—to support ongoing self-improvement and adaptation.

Within such environments, reinforcement learning from human or AI feedback on competitive performance \citep{christiano2017deep,bai2022constitutional} can enhance agents’ self-assessment and decision-making. Self-play \citep{silver2016mastering}, where agents compete against evolving versions of themselves, is especially effective for developing advanced strategies. General-purpose self-improvement methods \citep{zelikman2022star,wang2023voyager,singh2024beyond} can also be adapted, potentially via curriculum learning that introduces increasingly complex market scenarios. Despite their promise, these approaches face major challenges. Chief among them is the lack of robust benchmarks and environments that capture the real-world complexity and fluidity of competitive markets—far beyond what static matrix games offer. Equally critical is the development of evaluation metrics that can meaningfully assess the depth of an agent’s metacognitive and strategic reasoning.

\subsection{Market Design and Co-Evolution}
\label{subsec:market_design}

While this paper primarily focuses on the capabilities of LLM agents in future agentic labor markets, the design and governance of the labor markets they operate in are of paramount importance. Effective market design is crucial for ensuring market efficiency and fairness, thereby minimizing market failure like adverse selection and moral hazard \citep{akerlof1978market,spence1978job}, fostering trust and stability, building confidence among all participants (human principals and LLM agents alike), and encouraging broader adoption and long-term viability \citep{rothschild1978equilibrium,greif1993contract}. In the field of economics, careful market design, including mechanisms for information disclosure, signalling \citep{milgrom1981good}, and reputation systems \citep{dellarocas2003digitization} have been introduced to address adverse selection and moral hazard and will likely play a huge role in agentic markets too.

Notably, a dynamic feedback loop is likely to emerge: as LLM agents develop more sophisticated strategic reasoning, they may identify and exploit loopholes in existing market mechanisms. For instance, an agent might exploit moral hazard by subtly shirking on complex, hard-to-monitor tasks, delivering minimally compliant outputs to conserve resources. Alternatively, an agent could exploit adverse selection by strategically misrepresenting its specialized knowledge to secure tasks it is not suited for, leading to inefficient outcomes. This ``outsmarting'' necessitates the development of more adaptive and robust market designs, which, in turn, could drive agents to further their strategic capabilities, leading to a co-design process  between agents and market structures \citep{holland1992adaptation}.

This co-evolution underscores the critical need for adaptive market mechanisms that not only prevent exploitation but actively promote beneficial agent behaviors such as cooperation, honesty, and quality. This involves designing intelligent reward structures, dynamic reputation scoring, and transparent information flows that align individual agent utility with desirable market outcomes, potentially leveraging automated mechanism design to rapidly iterate and respond to new strategies \citep{roth2002economist,dellarocas2003digitization}. Furthermore, establishing ``circuit breakers'' or robust design principles is essential to stabilize this co-evolution. Addressing these multifaceted challenges effectively will necessitate strong cross-domain collaboration between machine learning researchers and economists.

\subsection{Safety, Alignment, and Governance}

In dynamic and complex market environments, traditional oversight mechanisms, that are often static, reactive, and manual, will rapidly become inadequate. They struggle to keep pace with the scale and complexity of agent interactions, and lack the sophistication to anticipate emergent behaviors in autonomous systems. Without more adaptive forms of oversight, autonomous agents risk developing unsafe behaviors \citep{zhuang2020consequences}, misaligned objectives \citep{russell2022human}, or exploiting reward functions through specification gaming \citep{krakovna2020specification}. These risks can materialize in the form of market destabilization, collusion, exploitation, resource monopolization, denial-of-service, or general behaviors that diverge from socially beneficial outcomes.

Mitigating safety and alignment risks requires a multi-layered approach, starting with market-level interventions. This includes designing labor markets that are adaptive and robust (\Cref{subsec:market_design}), and embedding safety constraints directly into their structure. These constraints may be technical (e.g., sandboxing, hard limits), economic (incentives and disincentives), or procedural (codes of conduct), all contributing to a broader “safety by design” paradigm where the environment itself reinforces aligned behavior. Equally important is effective governance: oversight processes—whether centralized, decentralized, or hybrid—must be capable of resolving disputes and adapting norms as conditions evolve. As adaptive market mechanisms could potentially become AI-driven, a deeper question arises: who governs the systems that govern the agents \citep{critch2020ai}?

The opaque and often proprietary nature of current agents makes it difficult to detect or interpret failures—particularly when developers control both the evaluation process and the disclosure of information. This underscores the need for independent, privacy-preserving auditing, including proactive certification \citep{emde2025shh} and adversarial evaluations such as red teaming \citep{perez2022red}. These efforts must be supported by advanced monitoring and interpretability tools that enable human or automated overseers to trace agent behavior, understand decision-making processes \citep{lindsey2025biology,ameisen2025circuit}, and intervene when necessary.

{\small
\bibliographystyle{unsrtnat}
\bibliography{bibfile}

\begin{thebibliography}{82}
\providecommand{\natexlab}[1]{#1}
\providecommand{\url}[1]{\texttt{#1}}
\expandafter\ifx\csname urlstyle\endcsname\relax
  \providecommand{\doi}[1]{doi: #1}\else
  \providecommand{\doi}{doi: \begingroup \urlstyle{rm}\Url}\fi

\bibitem[Lester et~al.(2019)Lester, Shourideh, Venkateswaran, and Zetlin-Jones]{lester2019screening}
Benjamin Lester, Ali Shourideh, Venky Venkateswaran, and Ariel Zetlin-Jones.
\newblock Screening and adverse selection in frictional markets.
\newblock \emph{Journal of Political Economy}, 127\penalty0 (1):\penalty0 338--377, 2019.

\bibitem[Holmstr{\"o}m(1979)]{holmstrom1979moral}
Bengt Holmstr{\"o}m.
\newblock Moral hazard and observability.
\newblock \emph{The Bell journal of economics}, pages 74--91, 1979.

\bibitem[Holmstr{\"o}m(1999)]{holmstrom1999managerial}
Bengt Holmstr{\"o}m.
\newblock Managerial incentive problems: A dynamic perspective.
\newblock \emph{The review of Economic studies}, 66\penalty0 (1):\penalty0 169--182, 1999.

\bibitem[Eloundou et~al.(2024)Eloundou, Manning, Mishkin, and Rock]{eloundou2024gpts}
Tyna Eloundou, Sam Manning, Pamela Mishkin, and Daniel Rock.
\newblock Gpts are gpts: Labor market impact potential of llms.
\newblock \emph{Science}, 384\penalty0 (6702):\penalty0 1306--1308, 2024.

\bibitem[Guo et~al.(2024)Guo, Chen, Wang, Chang, Pei, Chawla, Wiest, and Zhang]{guo2024large}
Taicheng Guo, Xiuying Chen, Yaqi Wang, Ruidi Chang, Shichao Pei, Nitesh~V Chawla, Olaf Wiest, and Xiangliang Zhang.
\newblock Large language model based multi-agents: a survey of progress and challenges.
\newblock In \emph{Proceedings of the Thirty-Third International Joint Conference on Artificial Intelligence}, pages 8048--8057, 2024.

\bibitem[Didolkar et~al.(2024)Didolkar, Goyal, Ke, Guo, Valko, Lillicrap, Rezende, Bengio, Mozer, and Arora]{didolkar2024metacognitive}
Aniket Didolkar, Anirudh Goyal, Nan~Rosemary Ke, Siyuan Guo, Michal Valko, Timothy Lillicrap, Danilo Rezende, Yoshua Bengio, Michael Mozer, and Sanjeev Arora.
\newblock Metacognitive capabilities of llms: An exploration in mathematical problem solving.
\newblock \emph{arXiv preprint arXiv:2405.12205}, 2024.

\bibitem[Zhang et~al.(2024)Zhang, Mao, Ge, Wang, Xia, Wu, Song, Lan, and Wei]{zhangllm}
Yadong Zhang, Shaoguang Mao, Tao Ge, Xun Wang, Yan Xia, Wenshan Wu, Ting Song, Man Lan, and Furu Wei.
\newblock Llm as a mastermind: A survey of strategic reasoning with large language models.
\newblock In \emph{First Conference on Language Modeling}, 2024.

\bibitem[Prato et~al.(2025)Prato, Huang, Parthasarathi, Sodhani, and Chandar]{prato2025large}
Gabriele Prato, Jerry Huang, Prasannna Parthasarathi, Shagun Sodhani, and Sarath Chandar.
\newblock Do large language models know how much they know?
\newblock \emph{arXiv preprint arXiv:2502.19573}, 2025.

\bibitem[Kadavath et~al.(2022)Kadavath, Conerly, Askell, Henighan, Drain, Perez, Schiefer, Hatfield-Dodds, DasSarma, Tran-Johnson, et~al.]{kadavath2022language}
Saurav Kadavath, Tom Conerly, Amanda Askell, Tom Henighan, Dawn Drain, Ethan Perez, Nicholas Schiefer, Zac Hatfield-Dodds, Nova DasSarma, Eli Tran-Johnson, et~al.
\newblock Language models (mostly) know what they know.
\newblock \emph{arXiv preprint arXiv:2207.05221}, 2022.

\bibitem[Bender et~al.(2021)Bender, Gebru, McMillan-Major, and Shmitchell]{bender2021dangers}
Emily~M Bender, Timnit Gebru, Angelina McMillan-Major, and Shmargaret Shmitchell.
\newblock On the dangers of stochastic parrots: Can language models be too big?
\newblock In \emph{Proceedings of the 2021 ACM conference on fairness, accountability, and transparency}, pages 610--623, 2021.

\bibitem[Li et~al.(2024)Li, Huang, Lin, Wu, Wan, and Sun]{li2024think}
Yuan Li, Yue Huang, Yuli Lin, Siyuan Wu, Yao Wan, and Lichao Sun.
\newblock I think, therefore i am: Awareness in large language models.
\newblock \emph{arXiv preprint arXiv:2401.17882}, 2024.

\bibitem[Steyvers and Peters(2025)]{steyvers2025metacognition}
Mark Steyvers and Megan~AK Peters.
\newblock Metacognition and uncertainty communication in humans and large language models.
\newblock \emph{arXiv preprint arXiv:2504.14045}, 2025.

\bibitem[Zhou et~al.(2025)Zhou, Jin, Shi, and Li]{zhou2025calibrating}
Ziang Zhou, Tianyuan Jin, Jieming Shi, and Qing Li.
\newblock Calibrating llm confidence with semantic steering: A multi-prompt aggregation framework.
\newblock \emph{arXiv preprint arXiv:2503.02863}, 2025.

\bibitem[Li et~al.(2025)Li, Li, Dong, Zhang, Zhang, Liu, Wang, Tang, and Liu]{li2025adaptive}
Wenjun Li, Dexun Li, Kuicai Dong, Cong Zhang, Hao Zhang, Weiwen Liu, Yasheng Wang, Ruiming Tang, and Yong Liu.
\newblock Adaptive tool use in large language models with meta-cognition trigger.
\newblock \emph{arXiv preprint arXiv:2502.12961}, 2025.

\bibitem[Haroon et~al.(2025)Haroon, Khan, Humayun, Gill, Amjad, Butt, Khan, and Gulzar]{haroon2025accurately}
Sabaat Haroon, Ahmad~Faraz Khan, Ahmad Humayun, Waris Gill, Abdul~Haddi Amjad, Ali~R Butt, Mohammad~Taha Khan, and Muhammad~Ali Gulzar.
\newblock How accurately do large language models understand code?
\newblock \emph{arXiv preprint arXiv:2504.04372}, 2025.

\bibitem[Hwang et~al.(2025)Hwang, Kim, Koo, Kang, Bae, and Jung]{hwang2025llms}
Yerin Hwang, Yongil Kim, Jahyun Koo, Taegwan Kang, Hyunkyung Bae, and Kyomin Jung.
\newblock Llms can be easily confused by instructional distractions.
\newblock \emph{arXiv preprint arXiv:2502.04362}, 2025.

\bibitem[Errica et~al.(2024)Errica, Siracusano, Sanvito, and Bifulco]{errica2024did}
Federico Errica, Giuseppe Siracusano, Davide Sanvito, and Roberto Bifulco.
\newblock What did i do wrong? quantifying llms' sensitivity and consistency to prompt engineering.
\newblock \emph{arXiv preprint arXiv:2406.12334}, 2024.

\bibitem[Kuhn et~al.(2022)Kuhn, Gal, and Farquhar]{kuhn2022clam}
Lorenz Kuhn, Yarin Gal, and Sebastian Farquhar.
\newblock Clam: Selective clarification for ambiguous questions with generative language models.
\newblock \emph{arXiv preprint arXiv:2212.07769}, 2022.

\bibitem[Sachdeva et~al.(2024)Sachdeva, Coleman, Kang, Ni, Hong, Chi, Caverlee, McAuley, and Cheng]{sachdeva2024train}
Noveen Sachdeva, Benjamin Coleman, Wang-Cheng Kang, Jianmo Ni, Lichan Hong, Ed~H Chi, James Caverlee, Julian McAuley, and Derek~Zhiyuan Cheng.
\newblock How to train data-efficient llms.
\newblock \emph{arXiv preprint arXiv:2402.09668}, 2024.

\bibitem[Wang et~al.(2023)Wang, Xie, Jiang, Mandlekar, Xiao, Zhu, Fan, and Anandkumar]{wang2023voyager}
Guanzhi Wang, Yuqi Xie, Yunfan Jiang, Ajay Mandlekar, Chaowei Xiao, Yuke Zhu, Linxi Fan, and Anima Anandkumar.
\newblock Voyager: An open-ended embodied agent with large language models.
\newblock \emph{arXiv preprint arXiv:2305.16291}, 2023.

\bibitem[Wu et~al.(2024)Wu, Han, Ding, Weng, Liu, Yao, Yu, and Kong]{wu2024copilot}
Zhiyong Wu, Chengcheng Han, Zichen Ding, Zhenmin Weng, Zhoumianze Liu, Shunyu Yao, Tao Yu, and Lingpeng Kong.
\newblock Os-copilot: Towards generalist computer agents with self-improvement.
\newblock \emph{arXiv preprint arXiv:2402.07456}, 2024.

\bibitem[Yao et~al.(2023)Yao, Zhao, Yu, Du, Shafran, Narasimhan, and Cao]{yao2023react}
Shunyu Yao, Jeffrey Zhao, Dian Yu, Nan Du, Izhak Shafran, Karthik Narasimhan, and Yuan Cao.
\newblock React: Synergizing reasoning and acting in language models.
\newblock In \emph{International Conference on Learning Representations (ICLR)}, 2023.

\bibitem[Wu et~al.(2025)Wu, Wang, Du, Jegelka, and Wang]{wu2025more}
Yuyang Wu, Yifei Wang, Tianqi Du, Stefanie Jegelka, and Yisen Wang.
\newblock When more is less: Understanding chain-of-thought length in llms.
\newblock \emph{arXiv preprint arXiv:2502.07266}, 2025.

\bibitem[Zheng et~al.(2024)Zheng, Mishra, Chen, Cheng, Chi, Le, and Zhou]{zheng2024take}
Huaixiu~Steven Zheng, Swaroop Mishra, Xinyun Chen, Heng-Tze Cheng, Ed~H. Chi, Quoc~V Le, and Denny Zhou.
\newblock Take a step back: Evoking reasoning via abstraction in large language models.
\newblock In \emph{The Twelfth International Conference on Learning Representations}, 2024.
\newblock URL \url{https://openreview.net/forum?id=3bq3jsvcQ1}.

\bibitem[van Duijn et~al.(2023)van Duijn, van Dijk, Kouwenhoven, de~Valk, Spruit, and van~der Putten]{van2023theory}
Max van Duijn, Bram van Dijk, Tom Kouwenhoven, Werner de~Valk, Marco Spruit, and Peter van~der Putten.
\newblock Theory of mind in large language models: Examining performance of 11 state-of-the-art models vs. children aged 7-10 on advanced tests.
\newblock In \emph{Proceedings of the 27th Conference on Computational Natural Language Learning (CoNLL)}, pages 389--402, 2023.

\bibitem[Strachan et~al.(2024)Strachan, Albergo, Borghini, Pansardi, Scaliti, Gupta, Saxena, Rufo, Panzeri, Manzi, et~al.]{strachan2024testing}
James~WA Strachan, Dalila Albergo, Giulia Borghini, Oriana Pansardi, Eugenio Scaliti, Saurabh Gupta, Krati Saxena, Alessandro Rufo, Stefano Panzeri, Guido Manzi, et~al.
\newblock Testing theory of mind in large language models and humans.
\newblock \emph{Nature Human Behaviour}, 8\penalty0 (7):\penalty0 1285--1295, 2024.

\bibitem[Zhu et~al.(2024)Zhu, Zhang, and Wang]{zhu2024language}
Wentao Zhu, Zhining Zhang, and Yizhou Wang.
\newblock Language models represent beliefs of self and others.
\newblock In \emph{Forty-first International Conference on Machine Learning}, 2024.
\newblock URL \url{https://openreview.net/forum?id=asJTE8EBjg}.

\bibitem[Sclar et~al.(2024)Sclar, Yu, Fazel-Zarandi, Tsvetkov, Bisk, Choi, and Celikyilmaz]{sclar2024explore}
Melanie Sclar, Jane Yu, Maryam Fazel-Zarandi, Yulia Tsvetkov, Yonatan Bisk, Yejin Choi, and Asli Celikyilmaz.
\newblock Explore theory of mind: Program-guided adversarial data generation for theory of mind reasoning.
\newblock \emph{arXiv preprint arXiv:2412.12175}, 2024.

\bibitem[Sarangi et~al.(2025)Sarangi, Elgarf, and Salam]{sarangi2025decompose}
Sneheel Sarangi, Maha Elgarf, and Hanan Salam.
\newblock Decompose-tom: Enhancing theory of mind reasoning in large language models through simulation and task decomposition.
\newblock In \emph{Proceedings of the 31st International Conference on Computational Linguistics}, pages 10228--10241, 2025.

\bibitem[Shapira et~al.(2024)Shapira, Levy, Alavi, Zhou, Choi, Goldberg, Sap, and Shwartz]{shapira2024clever}
Natalie Shapira, Mosh Levy, Seyed~Hossein Alavi, Xuhui Zhou, Yejin Choi, Yoav Goldberg, Maarten Sap, and Vered Shwartz.
\newblock Clever hans or neural theory of mind? stress testing social reasoning in large language models.
\newblock In \emph{Proceedings of the 18th Conference of the European Chapter of the Association for Computational Linguistics (Volume 1: Long Papers)}, pages 2257--2273, 2024.

\bibitem[Nguyen et~al.(2025)]{nguyen2025survey}
Hieu~Minh Nguyen et~al.
\newblock A survey of theory of mind in large language models: Evaluations, representations, and safety risks.
\newblock \emph{arXiv preprint arXiv:2502.06470}, 2025.

\bibitem[Gandhi et~al.(2023)Gandhi, Sadigh, and Goodman]{gandhi2023strategic}
Kanishk Gandhi, Dorsa Sadigh, and Noah~D. Goodman.
\newblock Strategic reasoning with language models, 2023.
\newblock arXiv preprint arXiv:2305.19165.

\bibitem[Akata et~al.(2025)Akata, Schulz, Coda-Forno, Oh, Bethge, and Schulz]{akata2025playing}
Elif Akata, Lion Schulz, Julian Coda-Forno, Seong~Joon Oh, Matthias Bethge, and Eric Schulz.
\newblock Playing repeated games with large language models.
\newblock \emph{Nature Human Behaviour}, pages 1--11, 2025.

\bibitem[Dagaev et~al.(2024)Dagaev, Paklina, and Parshakov]{dagaev2024strategizing}
Dmitry Dagaev, Sofiia Paklina, and Petr Parshakov.
\newblock Strategizing with ai: Insights from a beauty contest experiment.
\newblock \emph{Available at SSRN 4754435}, 2024.

\bibitem[Hua et~al.(2024)Hua, Liu, Li, Amayuelas, Chen, Jiang, Jin, Fan, Sun, Wang, et~al.]{hua2024game}
Wenyue Hua, Ollie Liu, Lingyao Li, Alfonso Amayuelas, Julie Chen, Lucas Jiang, Mingyu Jin, Lizhou Fan, Fei Sun, William Wang, et~al.
\newblock Game-theoretic llm: Agent workflow for negotiation games.
\newblock \emph{arXiv preprint arXiv:2411.05990}, 2024.

\bibitem[Duan et~al.(2024)Duan, Zhang, Diffenderfer, Kailkhura, Sun, Stengel-Eskin, Bansal, Chen, and Xu]{duan2024gtbench}
Jinhao Duan, Renming Zhang, James Diffenderfer, Bhavya Kailkhura, Lichao Sun, Elias Stengel-Eskin, Mohit Bansal, Tianlong Chen, and Kaidi Xu.
\newblock {GTB}ench: Uncovering the strategic reasoning capabilities of {LLM}s via game-theoretic evaluations.
\newblock In \emph{The Thirty-eighth Annual Conference on Neural Information Processing Systems}, 2024.
\newblock URL \url{https://openreview.net/forum?id=ypggxVWIv2}.

\bibitem[Lor\`{e} and Heydari(2024)]{lore2024strategicbehavior}
Nunzio Lor\`{e} and Babak Heydari.
\newblock Strategic behavior of large language models and the role of game structure versus contextual framing.
\newblock \emph{Scientific Reports}, 14\penalty0 (1):\penalty0 18490, 2024.

\bibitem[Bailis et~al.(2024)Bailis, Friedhoff, and Chen]{bailis2024werewolf}
Suma Bailis, Jane Friedhoff, and Feiyang Chen.
\newblock Werewolf arena: A case study in llm evaluation via social deduction.
\newblock \emph{arXiv preprint arXiv:2407.13943}, 2024.

\bibitem[Topsakal and Harper(2024)]{topsakal2024benchmarking}
Oguzhan Topsakal and Jackson~B Harper.
\newblock Benchmarking large language model (llm) performance for game playing via tic-tac-toe.
\newblock \emph{Electronics}, 13\penalty0 (8):\penalty0 1532, 2024.

\bibitem[FAIR(2022)]{meta2022human}
FAIR.
\newblock Human-level play in the game of diplomacy by combining language models with strategic reasoning.
\newblock \emph{Science}, 378\penalty0 (6624):\penalty0 1067--1074, 2022.

\bibitem[Jia et~al.(2025)]{jia2025evaluation}
Jingru Jia et~al.
\newblock Large language model strategic reasoning evaluation through behavioral game theory, 2025.
\newblock arXiv preprint arXiv:2502.20432.

\bibitem[Park et~al.(2023)Park, O'Brien, Cai, Morris, Liang, and Bernstein]{park2023generative}
Joon~Sung Park, Joseph O'Brien, Carrie~Jun Cai, Meredith~Ringel Morris, Percy Liang, and Michael~S Bernstein.
\newblock Generative agents: Interactive simulacra of human behavior.
\newblock In \emph{Proceedings of the 36th annual acm symposium on user interface software and technology}, pages 1--22, 2023.

\bibitem[Li et~al.(2023)Li, Yu, Li, Chen, and Khashanah]{li2023tradinggpt}
Yang Li, Yangyang Yu, Haohang Li, Zhi Chen, and Khaldoun Khashanah.
\newblock Tradinggpt: Multi-agent system with layered memory and distinct characters for enhanced financial trading performance.
\newblock \emph{arXiv preprint arXiv:2309.03736}, 2023.

\bibitem[Taubenfeld et~al.(2024)Taubenfeld, Dover, Reichart, and Goldstein]{taubenfeld2024systematic}
Amir Taubenfeld, Yaniv Dover, Roi Reichart, and Ariel Goldstein.
\newblock Systematic biases in llm simulations of debates.
\newblock In \emph{Proceedings of the 2024 Conference on Empirical Methods in Natural Language Processing}, pages 251--267, 2024.

\bibitem[Feng et~al.(2023)Feng, Luo, Wang, Tang, Yang, Shao, Mguni, Du, and Wang]{feng2023chessgpt}
Xidong Feng, Yicheng Luo, Ziyan Wang, Hongrui Tang, Mengyue Yang, Kun Shao, David Mguni, Yali Du, and Jun Wang.
\newblock Chessgpt: Bridging policy learning and language modeling.
\newblock \emph{Advances in Neural Information Processing Systems}, 36:\penalty0 7216--7262, 2023.

\bibitem[Wan et~al.(2024)Wan, Feng, Wen, McAleer, Wen, Zhang, and Wang]{wan2024alphazerolike}
Ziyu Wan, Xidong Feng, Muning Wen, Stephen~Marcus McAleer, Ying Wen, Weinan Zhang, and Jun Wang.
\newblock Alphazero-like tree-search can guide large language model decoding and training.
\newblock In \emph{Forty-first International Conference on Machine Learning}, 2024.
\newblock URL \url{https://openreview.net/forum?id=C4OpREezgj}.

\bibitem[Littman(1994)]{littman1994markov}
Michael~L Littman.
\newblock Markov games as a framework for multi-agent reinforcement learning.
\newblock In \emph{Machine learning proceedings 1994}, pages 157--163. Elsevier, 1994.

\bibitem[Vinyals et~al.(2019)Vinyals, Babuschkin, Czarnecki, Mathieu, Dudzik, Chung, Choi, Powell, Ewalds, Georgiev, et~al.]{vinyals2019grandmaster}
Oriol Vinyals, Igor Babuschkin, Wojciech~M Czarnecki, Micha{\"e}l Mathieu, Andrew Dudzik, Junyoung Chung, David~H Choi, Richard Powell, Timo Ewalds, Petko Georgiev, et~al.
\newblock Grandmaster level in starcraft ii using multi-agent reinforcement learning.
\newblock \emph{nature}, 575\penalty0 (7782):\penalty0 350--354, 2019.

\bibitem[Simon(1955)]{simon1955behavioral}
Herbert~A Simon.
\newblock A behavioral model of rational choice.
\newblock \emph{The quarterly journal of economics}, pages 99--118, 1955.

\bibitem[Parkes and Wellman(2015)]{parkes2015economic}
David~C Parkes and Michael~P Wellman.
\newblock Economic reasoning and artificial intelligence.
\newblock \emph{Science}, 349\penalty0 (6245):\penalty0 267--272, 2015.

\bibitem[Rahwan et~al.(2019)Rahwan, Cebrian, Obradovich, Bongard, Bonnefon, Breazeal, Crandall, Christakis, Couzin, Jackson, et~al.]{rahwan2019machine}
Iyad Rahwan, Manuel Cebrian, Nick Obradovich, Josh Bongard, Jean-Fran{\c{c}}ois Bonnefon, Cynthia Breazeal, Jacob~W Crandall, Nicholas~A Christakis, Iain~D Couzin, Matthew~O Jackson, et~al.
\newblock Machine behaviour.
\newblock \emph{Nature}, 568\penalty0 (7753):\penalty0 477--486, 2019.

\bibitem[Schaeffer et~al.(2023)Schaeffer, Miranda, and Koyejo]{schaeffer2023emergent}
Rylan Schaeffer, Brando Miranda, and Sanmi Koyejo.
\newblock Are emergent abilities of large language models a mirage?
\newblock \emph{Advances in Neural Information Processing Systems}, 36:\penalty0 55565--55581, 2023.

\bibitem[Du et~al.(2024)Du, Zeng, Dong, and Tang]{du2024understanding}
Zhengxiao Du, Aohan Zeng, Yuxiao Dong, and Jie Tang.
\newblock Understanding emergent abilities of language models from the loss perspective.
\newblock In \emph{The Thirty-eighth Annual Conference on Neural Information Processing Systems}, 2024.
\newblock URL \url{https://openreview.net/forum?id=35DAviqMFo}.

\bibitem[Lightman et~al.(2023)Lightman, Kosaraju, Burda, Edwards, Baker, Lee, Leike, Schulman, Sutskever, and Cobbe]{lightman2023let}
Hunter Lightman, Vineet Kosaraju, Yuri Burda, Harrison Edwards, Bowen Baker, Teddy Lee, Jan Leike, John Schulman, Ilya Sutskever, and Karl Cobbe.
\newblock Let's verify step by step.
\newblock In \emph{The Twelfth International Conference on Learning Representations}, 2023.

\bibitem[Costarelli et~al.(2024)Costarelli, Allen, Hauksson, Sodunke, Hariharan, Cheng, Li, Clymer, and Yadav]{costarelli2024gamebench}
Anthony Costarelli, Mat Allen, Roman Hauksson, Grace Sodunke, Suhas Hariharan, Carlson Cheng, Wenjie Li, Joshua Clymer, and Arjun Yadav.
\newblock Gamebench: Evaluating strategic reasoning abilities of llm agents.
\newblock \emph{arXiv preprint arXiv:2406.06613}, 2024.

\bibitem[Wang et~al.(2024)Wang, Feng, Li, Qin, Sui, and Kong]{wang2024tmgbench}
Haochuan Wang, Xiachong Feng, Lei Li, Zhanyue Qin, Dianbo Sui, and Lingpeng Kong.
\newblock Tmgbench: A systematic game benchmark for evaluating strategic reasoning abilities of llms.
\newblock \emph{arXiv preprint arXiv:2410.10479}, 2024.

\bibitem[Christiano et~al.(2017)Christiano, Leike, Brown, Martic, Legg, and Amodei]{christiano2017deep}
Paul~F Christiano, Jan Leike, Tom Brown, Miljan Martic, Shane Legg, and Dario Amodei.
\newblock Deep reinforcement learning from human preferences.
\newblock \emph{Advances in neural information processing systems}, 30, 2017.

\bibitem[Bai et~al.(2022)Bai, Kadavath, Kundu, Askell, Kernion, Jones, Chen, Goldie, Mirhoseini, McKinnon, et~al.]{bai2022constitutional}
Yuntao Bai, Saurav Kadavath, Sandipan Kundu, Amanda Askell, Jackson Kernion, Andy Jones, Anna Chen, Anna Goldie, Azalia Mirhoseini, Cameron McKinnon, et~al.
\newblock Constitutional ai: Harmlessness from ai feedback.
\newblock \emph{arXiv preprint arXiv:2212.08073}, 2022.

\bibitem[Silver et~al.(2016)Silver, Huang, Maddison, Guez, Sifre, Van Den~Driessche, Schrittwieser, Antonoglou, Panneershelvam, Lanctot, et~al.]{silver2016mastering}
David Silver, Aja Huang, Chris~J Maddison, Arthur Guez, Laurent Sifre, George Van Den~Driessche, Julian Schrittwieser, Ioannis Antonoglou, Veda Panneershelvam, Marc Lanctot, et~al.
\newblock Mastering the game of go with deep neural networks and tree search.
\newblock \emph{nature}, 529\penalty0 (7587):\penalty0 484--489, 2016.

\bibitem[Zelikman et~al.(2022)Zelikman, Wu, Mu, and Goodman]{zelikman2022star}
Eric Zelikman, Yuhuai Wu, Jesse Mu, and Noah Goodman.
\newblock Star: Bootstrapping reasoning with reasoning.
\newblock \emph{Advances in Neural Information Processing Systems}, 35:\penalty0 15476--15488, 2022.

\bibitem[Singh et~al.(2024)Singh, Co-Reyes, Agarwal, Anand, Patil, Garcia, Liu, Harrison, Lee, Xu, Parisi, Kumar, Alemi, Rizkowsky, Nova, Adlam, Bohnet, Elsayed, Sedghi, Mordatch, Simpson, Gur, Snoek, Pennington, Hron, Kenealy, Swersky, Mahajan, Culp, Xiao, Bileschi, Constant, Novak, Liu, Warkentin, Bansal, Dyer, Neyshabur, Sohl-Dickstein, and Fiedel]{singh2024beyond}
Avi Singh, John~D Co-Reyes, Rishabh Agarwal, Ankesh Anand, Piyush Patil, Xavier Garcia, Peter~J Liu, James Harrison, Jaehoon Lee, Kelvin Xu, Aaron~T Parisi, Abhishek Kumar, Alexander~A Alemi, Alex Rizkowsky, Azade Nova, Ben Adlam, Bernd Bohnet, Gamaleldin~Fathy Elsayed, Hanie Sedghi, Igor Mordatch, Isabelle Simpson, Izzeddin Gur, Jasper Snoek, Jeffrey Pennington, Jiri Hron, Kathleen Kenealy, Kevin Swersky, Kshiteej Mahajan, Laura~A Culp, Lechao Xiao, Maxwell Bileschi, Noah Constant, Roman Novak, Rosanne Liu, Tris Warkentin, Yamini Bansal, Ethan Dyer, Behnam Neyshabur, Jascha Sohl-Dickstein, and Noah Fiedel.
\newblock Beyond human data: Scaling self-training for problem-solving with language models.
\newblock \emph{Transactions on Machine Learning Research}, 2024.
\newblock ISSN 2835-8856.
\newblock URL \url{https://openreview.net/forum?id=lNAyUngGFK}.
\newblock Expert Certification.

\bibitem[Akerlof(1978)]{akerlof1978market}
George~A Akerlof.
\newblock The market for “lemons”: Quality uncertainty and the market mechanism.
\newblock In \emph{Uncertainty in economics}, pages 235--251. Elsevier, 1978.

\bibitem[Spence(1978)]{spence1978job}
Michael Spence.
\newblock Job market signaling.
\newblock In \emph{Uncertainty in economics}, pages 281--306. Elsevier, 1978.

\bibitem[Rothschild and Stiglitz(1978)]{rothschild1978equilibrium}
Michael Rothschild and Joseph Stiglitz.
\newblock Equilibrium in competitive insurance markets: An essay on the economics of imperfect information.
\newblock In \emph{Uncertainty in economics}, pages 257--280. Elsevier, 1978.

\bibitem[Greif(1993)]{greif1993contract}
Avner Greif.
\newblock Contract enforceability and economic institutions in early trade: The maghribi traders' coalition.
\newblock \emph{The American economic review}, pages 525--548, 1993.

\bibitem[Milgrom(1981)]{milgrom1981good}
Paul~R Milgrom.
\newblock Good news and bad news: Representation theorems and applications.
\newblock \emph{The Bell Journal of Economics}, pages 380--391, 1981.

\bibitem[Dellarocas(2003)]{dellarocas2003digitization}
Chrysanthos Dellarocas.
\newblock The digitization of word of mouth: Promise and challenges of online feedback mechanisms.
\newblock \emph{Management science}, 49\penalty0 (10):\penalty0 1407--1424, 2003.

\bibitem[Holland(1992)]{holland1992adaptation}
John~H Holland.
\newblock \emph{Adaptation in natural and artificial systems: an introductory analysis with applications to biology, control, and artificial intelligence}.
\newblock MIT press, 1992.

\bibitem[Roth(2002)]{roth2002economist}
Alvin~E Roth.
\newblock The economist as engineer: Game theory, experimentation, and computation as tools for design economics.
\newblock \emph{Econometrica}, 70\penalty0 (4):\penalty0 1341--1378, 2002.

\bibitem[Zhuang and Hadfield-Menell(2020)]{zhuang2020consequences}
Simon Zhuang and Dylan Hadfield-Menell.
\newblock Consequences of misaligned ai.
\newblock \emph{Advances in Neural Information Processing Systems}, 33:\penalty0 15763--15773, 2020.

\bibitem[Russell(2022)]{russell2022human}
Stuart Russell.
\newblock Human-compatible artificial intelligence., 2022.

\bibitem[Krakovna et~al.(2020)Krakovna, Uesato, Mikulik, Rahtz, Everitt, Kumar, Kenton, Leike, and Legg]{krakovna2020specification}
Victoria Krakovna, Jonathan Uesato, Vladimir Mikulik, Matthew Rahtz, Tom Everitt, Ramana Kumar, Zac Kenton, Jan Leike, and Shane Legg.
\newblock Specification gaming: the flip side of ai ingenuity.
\newblock \emph{DeepMind Blog}, 3, 2020.

\bibitem[Critch and Krueger(2020)]{critch2020ai}
Andrew Critch and David Krueger.
\newblock Ai research considerations for human existential safety (arches).
\newblock \emph{arXiv preprint arXiv:2006.04948}, 2020.

\bibitem[Emde et~al.(2025)Emde, Paren, Arvind, Kayser, Rainforth, Lukasiewicz, Ghanem, Torr, and Bibi]{emde2025shh}
Cornelius Emde, Alasdair Paren, Preetham Arvind, Maxime Kayser, Tom Rainforth, Thomas Lukasiewicz, Bernard Ghanem, Philip~HS Torr, and Adel Bibi.
\newblock Shh, don't say that! domain certification in llms.
\newblock \emph{arXiv preprint arXiv:2502.19320}, 2025.

\bibitem[Perez et~al.(2022)Perez, Huang, Song, Cai, Ring, Aslanides, Glaese, McAleese, and Irving]{perez2022red}
Ethan Perez, Saffron Huang, Francis Song, Trevor Cai, Roman Ring, John Aslanides, Amelia Glaese, Nat McAleese, and Geoffrey Irving.
\newblock Red teaming language models with language models.
\newblock In \emph{Proceedings of the 2022 Conference on Empirical Methods in Natural Language Processing}, pages 3419--3448, 2022.

\bibitem[Lindsey et~al.(2025)Lindsey, Gurnee, Ameisen, Chen, Pearce, Turner, Citro, Abrahams, Carter, Hosmer, Marcus, Sklar, Templeton, Bricken, McDougall, Cunningham, Henighan, Jermyn, Jones, Persic, Qi, Thompson, Zimmerman, Rivoire, Conerly, Olah, and Batson]{lindsey2025biology}
Jack Lindsey, Wes Gurnee, Emmanuel Ameisen, Brian Chen, Adam Pearce, Nicholas~L. Turner, Craig Citro, David Abrahams, Shan Carter, Basil Hosmer, Jonathan Marcus, Michael Sklar, Adly Templeton, Trenton Bricken, Callum McDougall, Hoagy Cunningham, Thomas Henighan, Adam Jermyn, Andy Jones, Andrew Persic, Zhenyi Qi, T.~Ben Thompson, Sam Zimmerman, Kelley Rivoire, Thomas Conerly, Chris Olah, and Joshua Batson.
\newblock On the biology of a large language model.
\newblock \emph{Transformer Circuits Thread}, 2025.
\newblock URL \url{https://transformer-circuits.pub/2025/attribution-graphs/biology.html}.

\bibitem[Ameisen et~al.(2025)Ameisen, Lindsey, Pearce, Gurnee, Turner, Chen, Citro, Abrahams, Carter, Hosmer, Marcus, Sklar, Templeton, Bricken, McDougall, Cunningham, Henighan, Jermyn, Jones, Persic, Qi, Ben~Thompson, Zimmerman, Rivoire, Conerly, Olah, and Batson]{ameisen2025circuit}
Emmanuel Ameisen, Jack Lindsey, Adam Pearce, Wes Gurnee, Nicholas~L. Turner, Brian Chen, Craig Citro, David Abrahams, Shan Carter, Basil Hosmer, Jonathan Marcus, Michael Sklar, Adly Templeton, Trenton Bricken, Callum McDougall, Hoagy Cunningham, Thomas Henighan, Adam Jermyn, Andy Jones, Andrew Persic, Zhenyi Qi, T.~Ben~Thompson, Sam Zimmerman, Kelley Rivoire, Thomas Conerly, Chris Olah, and Joshua Batson.
\newblock Circuit tracing: Revealing computational graphs in language models.
\newblock \emph{Transformer Circuits Thread}, 2025.
\newblock URL \url{https://transformer-circuits.pub/2025/attribution-graphs/methods.html}.

\bibitem[Daly(1974)]{daly1974economics}
Herman~E Daly.
\newblock The economics of the steady state.
\newblock \emph{The american economic review}, 64\penalty0 (2):\penalty0 15--21, 1974.

\bibitem[Maskin and Tirole(2001)]{maskin2001markov}
Eric Maskin and Jean Tirole.
\newblock Markov perfect equilibrium: I. observable actions.
\newblock \emph{Journal of Economic Theory}, 100\penalty0 (2):\penalty0 191--219, 2001.

\bibitem[Belkin et~al.(2013)Belkin, Kurtzberg, and Naquin]{belkin2013signaling}
Liuba~Y. Belkin, Terri~R. Kurtzberg, and Charles~E. Naquin.
\newblock Signaling dominance in online negotiations: The role of affective tone.
\newblock \emph{Negotiation and Conflict Management Research}, 6\penalty0 (4):\penalty0 285--304, 2013.

\bibitem[Thaler(2016)]{thaler2016behavioral}
Richard~H Thaler.
\newblock Behavioral economics: Past, present, and future.
\newblock \emph{American economic review}, 106\penalty0 (7):\penalty0 1577--1600, 2016.

\bibitem[Chen et~al.(2015)Chen, Micali, and Pass]{chen2015tight}
Jing Chen, Silvio Micali, and Rafael Pass.
\newblock Tight revenue bounds with possibilistic beliefs and level-k rationality.
\newblock \emph{Econometrica}, 83\penalty0 (4):\penalty0 1619--1639, 2015.

\end{thebibliography}
}
\clearpage
\appendix
\section{Current Contracts for LLMs vs. Future Contracts for Agents}
Future contracts for AI agents will differ markedly from the current contracts. For LLMs, two predominant types of contracts currently exist: a pay per use model based on the amount of input/output tokens used, and subscriptions with a fixed monthly fee. These contracts put a heavy burden on the client to ensure that the AI will provide value. By contrast, future labor markets are sure to involve more varied and sophisticated contracts. For instance, an agent could be paid for achieving certain milestones instead of a flat rate for usage. A lawyer agent could be hired and paid only if it is able to win a lawsuit, and a sales agent could be hired and paid only after achieving a certain number of sales. While less risky for the client, such contracts put a greater burden on the AI agent. As such, AI agents receiving these contracts will need more advanced reasoning capabilities to evaluate them.

\section{Agentic Moral Hazard Concerns}
In this subsection, we discuss several new moral hazard concerns that are unique to AI agents. For instance, AI agents have the incentive to learn as much information as possible from its clients so that it can train itself and improve its own abilities for future jobs. As such, data privacy can be a major concern for clients of AI agents, and this concern is much more severe than currently with LLMs. AI agents will have access to tool use and thus have much more power over the information that they are able to access. Another avenue for moral hazard is through misaligned AI agents, as safety training for AI agents can be much more difficult than for LLMs, and their malicious activities will be much more difficult to detect.

More subtle forms of moral hazard are also present for AI agents. Currently, there are only a few research labs that are producing frontier AI models, and this situation could plausibly continue well into the future given the high training and compute costs associated with frontier AI models. This leads to a lack of options in where clients can source their AI agents. In competitive environments, such as two companies competing on sales of similar products, there could be conflicts of interest if both companies source their AI agents from the same research lab.\footnote{In economic terms this is known as an oligopoly, as the supply of labor (AI agents) is controlled by only a few research labs.} For instance, one of the companies could attempt to bribe the AI provider to sabotage the performance of the AI agents used by their competitors. A sufficiently intelligent and profit-maximizing AI agent could even attempt to extract a bribe from one of the companies autonomously if not properly aligned.\footnote{Such concerns are highly prevalent in current labor markets. For instance, it is a common practice that two sides of a legal dispute do not seek representation from the same law firm due to conflict of interest concerns.}

\section{Hiring Other Agents}

Another important aspect of strategic decision-making is the ability of an agent to hire other agents (human or AI) on the labor market to assist with tasks. Thus, the agent participates in both the supply and demand sides of the market. We call these hired agents \emph{colleagues}, denoting their types by \(\{\hat{\theta}_{k}\}\) and their reputations by \(\{\hat{a}_{k}\}\). Naturally, the agent’s competitors can hire their own colleagues as well. Formally, we now allow the reward function to depend on all of these parties:
\(
Z_t^{k}\!\bigl(
\theta_i,a_i,\{\hat{\theta}_{k},\hat{a}_{k}\},
\{\theta_{-i},a_{-i}\},
\{\hat{\theta}_{-k},\hat{a}_{-k}\}
\bigr).
\) As a concrete example, a lawyer‑agent might hire another agent with medical expertise for a medical malpractice case. This mirrors how law firms consult expert witnesses. The key point is that agents themselves may lack expertise in every domain, which is a significant departure from the generalist LLM systems of today.

One simple example (ignoring competition and assuming just one colleague) is
\[
Z_t^{k}\!\bigl(\theta_i,a_i,\hat{\theta}_{k},\hat{a}_{k}\bigr)
=\bigl(\theta_{i,t}+a_{i,t}+\varepsilon_{i,t}\bigr)
 \bigl(\hat{\theta}_{k}+\hat{a}_{k}+\varepsilon_{k,t}\bigr).
\]

Here, output equals the product of the agent’s own contribution and that of its colleague. High output by a colleague acts as a force multiplier for the agent’s own output, whereas poor output by a colleague drags down the agent’s output quality. With such a functional form, the agent’s choice of colleague and the colleague’s action becomes very important for the success of the project.

Allowing an agent to hire other agents greatly increases reasoning demands. The agent must contend with adverse selection, moral hazard, and reputation when making hiring decisions. Because the types \(\hat{\theta}_{k}\) of potential colleagues are not fully observable (adverse selection), the agent must use their reputation to estimate their true quality. The agent also may not be able to monitor the actions taken by its colleagues (moral hazard), and so whatever contract it offers to its colleague must incentivize sufficient effort. Moreover, the agent’s colleagues, competitors, and colleagues of competitors also all have their own metacognition and strategic reasoning, and the agent must therefore anticipate and reason about the reasoning of others.

\section{Case Studies}
In this section, we present case studies to highlight the agent reasoning that will be required in several types of jobs.

\textbf{Lawyer} In the current labor market, popular types of contracts for lawyers include fixed rate, pay for hours worked (billable hours), and pay contingent upon winning a case. AI lawyer agents will need advanced reasoning abilities when deciding whether to accept such contracts. Especially for jobs that pay contingent upon winning a case, AI agents must be able to reason about their ability to win the lawsuit before agreeing to accept a job. This is a very complicated problem, which involves reasoning about the abilities of the potential opposition, which may be a mixture of humans and AI. Agents will need to have the ability to synthesize information from online and offline sources about the merits of a case and the expected strength/capabilities of the opposition in a case. Agents with previous experience in legal cases must be able to draw upon that experience and make improvements as needed to their algorithms. Reputation can be important as well, as agents with higher reputations could command higher fees from clients. As such, it could be beneficial for agents to initially work for lower wages to increase their reputation and then get higher wages in the future.

\textbf{Scientific Researcher} Scientific research is a highly competitive field, with success criteria including the ability to research and publish major results, ability to attract high-quality collaborators, and ability to gain funding for laboratories and equipment. An AI agent must be able to reason about the difficulty of the research problem, capabilities of collaborators, funding and resources that will be provided, and rival research teams and their abilities. Moral hazard is an issue, as researchers that are not monitored have an incentive to magnify or falsify their results. Agents will need to be appropriately aligned in order to not falsify or hallucinate their research output. Reputation is important to obtain funding, which in turn allows for more experiments and thus more opportunities for success. As such, it will be important for agents to build their reputations to gain such funding. Researcher agents will also often need to work with capable collaborators and must strategically reason about whether to partner with specific humans or AI agents on their projects.

\textbf{Writer} Writer is a very saturated field, and is  often thought of as a “superstar” industry, where a few well-known individuals can make significantly more money than the average writer. To determine whether it can generate revenue, an AI writer must therefore reason about its writing abilities in comparison to its competition. This involves using metacognitive and strategic reasoning to determine whether it has any unique characteristics that would allow it to stand out. The AI agent must also understand the tastes of its potential clients (i.e. readership) and whether it can achieve their requests. Developing a strong reputation is very important in this field, and as such the AI agent needs to have a long-term plan of how to build its reputation over time, perhaps by offering its initial works for low costs. An AI writer would also need to reason about the best ways to improve its abilities over time and adjust to changing tastes in the population towards books. Finding collaborators in writing is typically less important, but an agent may need to hire a marketing agent to help it publicize its writing.

\section{Overview of Current State on Metacognition and Strategic Reasoning}

This appendix provides a high-level summary of the current state and key open questions of the two core capabilities discussed in the main text: metacognition and strategic reasoning. Each encompasses several components—self-assessment, task understanding, and strategy evaluation for metacognition; theory of mind, strategic decision-making, and strategic learning for strategic reasoning. \Cref{tab:metacognition_overview,tab:strategic_overview} synthesize existing evidence and identify critical gaps that must be addressed to enable LLM agents to operate effectively in future agentic labor markets.

\begin{table}[h!]
\centering
\caption{\textbf{Overview of metacognition capabilities.}}
\label{tab:metacognition_overview}
\renewcommand{\arraystretch}{0.5}
\begin{adjustbox}{max width=\textwidth}
\begin{tabular}{>{\raggedright\arraybackslash}p{1.9cm} >{\raggedright\arraybackslash}p{7cm} >{\raggedright\arraybackslash}p{7cm}}
\toprule
\textbf{Component} & \textbf{Current capabilities} & \textbf{Open questions} \\
\midrule
\textbf{Understanding of self} 
& 
\vspace{-0.65em}
\begin{itemize}[left=0pt, nosep]
  \item Emerging awareness of knowledge and capability boundaries
  \item Calibration can be improved with prompting, finetuning, and scale
\end{itemize}
& 
\vspace{-0.65em}
\begin{itemize}[left=0pt, nosep]
  \item Lacks robustness and generalization on OOD tasks
  \item Bridging the gap between verbalized and implicit uncertainty
\end{itemize}
\\
\textbf{Understanding of tasks} &
\vspace{-0.65em}
\begin{itemize}[left=0pt, nosep]
  \item Awareness of task difficulty relative to capability
  \item Construct continual learning curriculum based on task understanding
\end{itemize} &
\vspace{-0.65em}
\begin{itemize}[left=0pt, nosep]
  \item Brittle dependence on syntax rather than semantics
  \item Consistent and robust assessment on ambiguous or ill-defined tasks
\end{itemize}
\\
\textbf{Understanding of strategies} 
\vspace{-0.65em}
& 
\vspace{-0.65em}
\begin{itemize}[left=0pt, nosep]
  \item Dynamic adaptation of problem-solving strategies given feedback
  \item Specialized prompting and training improves evaluation robustness
\end{itemize} 
\vspace{-0.65em}
&
\vspace{-0.65em}
\begin{itemize}[left=0pt, nosep]
  \item Enabling autonomous adaptation beyond explicit prompt scaffolding
  \item Generalization of dynamic strategy selection under-investgiated in open-ended domains
\end{itemize}
\vspace{-0.65em}
\\
\bottomrule
\end{tabular}
\end{adjustbox}
\end{table}

\begin{table}[h!]
\centering
\caption{\textbf{Overview of strategic reasoning capabilities.}}
\label{tab:strategic_overview}
\renewcommand{\arraystretch}{0.5}
\begin{adjustbox}{max width=\textwidth}
\begin{tabular}{>{\raggedright\arraybackslash}p{1.9cm} >{\raggedright\arraybackslash}p{6.5cm} >{\raggedright\arraybackslash}p{6.5cm}}
\toprule
\textbf{Component} & \textbf{Current capabilities} & \textbf{Open questions} \\
\midrule
\textbf{Theory of Mind (ToM)} & 
\vspace{-0.65em}
\begin{itemize}[left=0pt, nosep]
  \item Emerging mental state inference (e.g., false belief, irony)
  \item Internal representations of belief states identified
  \item Performance improvements by scaling, prompting, fine-tuning
\end{itemize}
\vspace{-0.65em}
& 
\vspace{-0.65em}
\begin{itemize}[left=0pt, nosep]
  \item Lacks robustness, fails on adversarial/varied tasks
  \item Struggles on complex ToM
  \item Benchmark limitations, potential overfitting
  \item Weak state tracking, inconsistent inference commitment
\end{itemize}
\vspace{-0.65em}
\\
\textbf{Strategic decision-making} 
\vspace{-0.65em}
&
\vspace{-0.65em}
\begin{itemize}[left=0pt, nosep]
  \item Human-like participation/strategies in game-theoretic scenarios
  \item Adaptation in repeated games
  \item Can formalize real-world scenarios into game models
\end{itemize} 
\vspace{-0.65em}
&
\vspace{-0.65em}
\begin{itemize}[left=0pt, nosep]
  \item Sub-optimal play in complete, deterministic matrix games
  \item Struggles in more complex payoff matrices, deeper sequential trees
  \item Behavioral biases from framing, payoffs, identity and emotional bias
\end{itemize}
\vspace{-0.65em}
\\
\textbf{Strategic learning} 
\vspace{-0.65em}
& 
\vspace{-0.65em}
\begin{itemize}[left=0pt, nosep]
  \item Strategy adaptation from real-time feedback/opponent modeling
  \item Demonstrated feasibility in two-player settings
\end{itemize} 
\vspace{-0.65em}
&
\vspace{-0.65em}
\begin{itemize}[left=0pt, nosep]
  \item Struggle with multi-agent, extensive form games with explicit utility maximization
\end{itemize}
\vspace{-0.65em}
\\

\bottomrule
\end{tabular}
\end{adjustbox}
\end{table}

\section{Comparison of Metacognitive and Strategic Reasoning with Economic Models of Learning and Reasoning}

Although the mathematical formulation presented in this paper is similar to the formulations in economic models, the reasoning of AI agents will function very differently from how agents act in economic models. We describe the main differences in this section. We start by providing the reader with an overview of how reasoning functions in typical economic models. 

\subsection{Overview of Economics Learning and Reasoning}

\textbf{Economic models} usually assume \textit{fully rational} agents who take actions according to a concept known as \textit{Nash Equilibrium}, whereby agents do not want to deviate from their own chosen actions given what their opponents have chosen. As we will discuss, these assumptions impose \textit{strong restrictions} on the behavior of the agents in the models. The overall impact is to greatly \textit{limit the complexity} of economics models and the situations they can assess. 

\textbf{Full rationality} assumes that agents will make the optimal decisions for themselves given their current information, and a Nash equilibrium is the resulting outcome when all agents make optimal decisions and none wishes to deviate. Importantly, these concepts imply that agents have beliefs about their opponents actions and strategies, that all opponents of the agent also act rationally, and that the beliefs of the opponents about the agent are also correct. These concepts are more reasonable for simple types of games, such as matrix games in which players take discrete actions. For instance, consider the game rock-paper-scissors. The unique Nash equilibrium of this game is when both players randomize among each action with probability 1/3. As neither player can gain an advantage by deviating to a different strategy, both players are acting optimally and hold accurate beliefs about their opponent's actions. Such a strategy is simple and intuitive enough for most human players to understand. 

To prove results under the assumption of fully rational agents, economic models are thus deliberately designed to be \textit{relatively simple}, and they often make strong assumptions about functional forms or model parameters to prove cleaner results. Models usually ignore most real-world frictions, but they instead focus selectively on a specific topic of interest. Adverse selection and moral hazard are often considered separately in economic models to highlight the individual impact of each force. The models of adverse selection typically assume assumptions such as the “single-crossing property” on agent preferences based on their types to prove their results.\footnote{This is known in economics as the Spence-Mirrlees condition.}  Models of moral hazard likewise often make strong assumptions on the cost functions of agents to characterize the ensuing contracts.

\textbf{Learning in Economics.} Economic models that include \textit{learning} also tend to make simplifying and strong assumptions about the environment in order to prove clean results. As economics assumes that agents are fully rational and thus already make optimal decisions, learning does not entail improvements to the agent's reasoning capabilities. Instead, agent learning typically is limited to learning about or improvements to the agent's type, in a similar vein as we presented in our paper. For instance, economic models may assume that agents can invest in a costly action that is able to improve their type in the next period with some probability. In addition, other parts of the system are typically kept relatively simple in models of agent learning, in order for the optimal learning policy to be tractable. As such, the extent of competition with other agents or changes to the system over time are usually kept limited.

\textbf{Dynamic Economic Models.} Economic models with multiple or infinite periods usually consider a \textit{steady state} \citet{daly1974economics}, whereby key elements of the system are constant over time. Of note, such steady states do not necessarily need to assume that the entire system as a whole is unchanging, but merely that inflows and outflows into the system are balanced. The related concept of a \textit{Markov Perfect Equilibrium} \citep{maskin2001markov} is also commonly used in the economic theory literature to restrict the scope of actions and beliefs. It assumes that all agents have strategies that do not depend on the past outside of a predefined "state" variable, which could for instance represent the expected belief about an agent's quality in models with reputation. As such, the assumption eliminates the need to consider any path-dependent aspects of the system. This serves to greatly reduce the scale of the action and belief space for agents in the model, which can significantly improve tractability at the expense of realism. Notably, both concepts of \textit{steady states} and \textit{Markov Perfect Equilibrium} typically preclude the possibility of significant changes to the system over time, such as the entry of new types of agents or the introduction of new technologies. Such factors would introduce a level of path-dependence and complexity that these frameworks would be unable to capture.

\textbf{Role of Language.} We note that another significant difference is the lack of natural language in economic models, whereas AI agents are large language models. Economics models typically abstract away from natural language, as they consider it a form of “cheap talk” that does not affect payoffs. The concepts of Nash equilibrium and rationality don’t typically consider the exact language used in communications, so for instance a message conveyed in a “warm” vs. “cold” tone has no direct impact. LLMs by contrast are very dependent on language order and phrasing, as slight changes in prompts can lead to drastic changes in output. The reasoning of AI agents, and their communications, will be very much influenced by natural language, unlike in economics. This makes their reasoning more natural and “human-like” than economics models, as the tone of language has been found to be very important in real-world studies. For instance, the field of negotiation theory for instance has found that the tone of language is very important for negotiation success.\footnote{See for instance \citet{belkin2013signaling}.}

\subsection{Comparison with Metacognitive and Strategic Reasoning}

\textbf{Our Labor Market Framework.} By contrast, the reasoning AI agents will use in our labor market scenario involves very a complex model that combines many economic forces. We presented a \textit{non-stationary} and \textit{dynamic} process, and agents are taking actions in a \textit{non-steady state} setting. AI agents will not act according to full rationality. In any case the model that is considered is far too complex to explicitly compute an optimal solution. Instead, AI agents will make decisions following the specific training process that was used, and they can learn over time based on their past experiences as well. Metacognition affects the agent's ability to understand and improve its own reasoning processes over time, and agents with superior metacognitive reasoning will demonstrate better performance on such tasks.

For the more complicated aspects of our framework, such as entry of new competitors or changes in technology, agents may make very suboptimal decisions initially and need to improve their reasoning over time based on their past experiences. In addition, agents may have limited understanding about their opponents, and their beliefs about opponent strategies may be far from accurate. As such, an agent will also need to improve its strategic reasoning skills over time, in order to better predict the actions of its competitors. In this way, it is also critical for the agent to have strong strategic reasoning capabilities.

\textbf{Modelling Difficulty.} Freeing AI agents from the constraints of full rationality and Nash equilibrium makes the analysis of some situations easier, while others become more complicated. For very complex scenarios, where computing the optimal action can be impossible, AI agents can instead use heuristics that are closer to actual human behavior. This can greatly simplify the reasoning involved, allowing AI agents to use a “good enough” solution instead of forcing them to solve for an intractable optimal solution. Analysis of such models would thus focus on the expected end result of agents following heuristics. 

On the other hand, for simple games, such as the rock-paper-scissors, the concept of Nash equilibrium simplifies the problem. If an AI agent knows that their opponent will be taking a Nash equilibrium strategy and purposely mixing among all actions with probability 1/3, then it doesn’t matter what the AI agent does as the opponent’s strategy is impossible to exploit. However, if the AI agent’s opponent is not taking the Nash equilibrium action, then the AI agent may need to reason significantly to predict what their opponent will do and the best way to exploit it.

Our labor market framework falls into the class of "complicated" problems, and as such analysis of systems with agents that follow heuristics is likely simpler than considering fully rational behavior, as computing optimal strategies may be infeasible. Analysis of our labor market framework would involve deep dives into the metacognitive and strategic reasoning abilities of agents, with experiments and simulators designed to test how the labor market evolves as agentic reasoning abilities are improved over time and new agents enter the system.

\subsection{Comparison with Behavioral Economics}

\textbf{Behavioral Economics}. Even for very simple games, like the well-known prisoner’s dilemma game (a two-action matrix game), humans often do not behave according to Nash Equilibrium. Indeed, the field of experimental economics has found many deviations from Nash Equilibrium behavior when humans are asked to participate in experiments. These findings gave rise to a subfield of economics known as \textit{behavioral economics}, which tries to understand the myriad ways in which humans deviate from fully rational behavior. 

Behavioral economics accounts for common fallacies in human reasoning, such as overconfidence and loss aversion \citep{thaler2016behavioral}. A major portion of behavioral economics is dedicated to assessing non-fully rational modes of thinking. The assumption of full rationality is replaced by \textit{bounded rationality}, in which agents make suboptimal decisions by following along with \textit{heuristics} instead of computing the optimal choice. One example is \textit{level-k reasoning}, which assumes that some agents may only optimize against very basic strategies instead of the full set of complex strategies their opponents have available. Research in this area then finds the potential outcomes and implications of this type of reduced rationality for agents \citep{chen2015tight}.

\textbf{The Need for ``Agentic" Behavioral Economics}. The reasoning and learning performed by AI agents is much closer to behavioral economics than traditional economics. Agents do not act fully rationally, as they do not explicitly compute the optimal solutions to their value functions or accurately model the actions and beliefs of their opponents. Instead, agents tend to follow heuristics, which may be imprinted on them during their training phase. For instance, if an agent was trained on many examples of loss aversion, then the agent itself may exhibit loss aversion as well.

However, a key difference between agent reasoning and behavioral economics is that agent reasoning is much more \textit{malleable}, and potential errors or problems in its reasoning can be mitigated through an improved training process. If an agent exhibits too much loss aversion, then the next training run could try to counteract this by explicitly including less loss aversion samples in the training data. This methodology follows closely with current work on agent \textit{safety and alignment}, which tries to ensure that agents do not take malicious actions. A similar type of process may be needed to eliminate \textit{reasoning errors}. Of note, given that current LLMs are trained on mostly human generated data, it is likely that LLMs exhibit similar cognitive biases as humans. But this could be attenuated in the future as greater amounts of \textit{synthetic data} are used in training agents. Synthetic data may be free from the cognitive biases of humans, although new AI-specific cognitive biases could arise.

We believe there is a need for an \textit{agentic behavioral economics} that studies how agents learn and reason in the real-world. This line of research could use similar methods as in behavioral economics, such as experimentation. With agents, the experimentation process could be much faster, as agents could be put into simulators that allow them to interact and compete against each other over many rounds. Research would then assess what types of "agentic" cognitive biases arise over time from these simulations, and compare it against known cognitive biases found in humans.
\end{document}